\documentclass[11pt,a4paper]{article}
\usepackage[hyperref]{acl2019}
\usepackage{times}
\usepackage{latexsym}
\usepackage{graphicx}
\graphicspath{ {./images/} }
\usepackage{textcomp}
\usepackage{upquote}
\usepackage{enumitem}
\usepackage[T1]{fontenc}
\usepackage{url}

\aclfinalcopy 
\def\aclpaperid{37} 

\title{Sanskrit Segmentation Revisited}

\author{Sriram Krishnan and Amba Kulkarni \\
  Department of Sanskrit Studies \\
  University of Hyderabad \\
  \texttt{sriramk8@gmail.com, apksh.uoh@nic.in} \\}

\date{}

\begin{document}
\maketitle
\begin{abstract}
Computationally analyzing Sanskrit texts requires proper segmentation in the initial stages. There have been various tools developed for Sanskrit text segmentation. Of these, G\'erard Huet's Reader in the Sanskrit Heritage Engine analyzes the input text and segments it based on the word parameters - phases like iic, ifc, Pr, Subst, etc., and \textit{sandhi} (or transition) that takes place at the end of a word with the initial part of the next word. And it enlists all the possible solutions differentiating them with the help of the phases. The phases and their analyses have their use in the domain of sentential parsers. In segmentation, though, they are not used beyond deciding whether the words formed with the phases are morphologically valid. This paper tries to modify the above segmenter by ignoring the phase details (except for a few cases), and also proposes a probability function to prioritize the list of solutions to bring up the most valid solutions at the top.
\end{abstract}

\section{Introduction} \label{sec:introduction}

Every Sanskrit sentence in the \textit{sa\d{m}hit\={a}} form (continuous sandhied text) is required to be segmented into proper morphologically acceptable words and the obtained result should agree with syntactic and semantic correctness for it's proper understanding. The obtained segmented text consists of individual words where even the compounds are segmented into their components. And there can be more than one segmentation for the same \textit{sa\d{m}hit\={a}} text. The segmented form does not provide any difference in the sense of the text when compared with the \textit{sa\d{m}hit\={a}} form except for the difference in the phonology of the words where it can be observed that the end part of the initial word together with the first letter of the next word undergoes phonetic change. The \textit{sa\d{m}hit\={a}} form, in fact, represents the text similar to a speech text because the knowledge transfer, in the olden days, was predominantly based on oral rendition. But now, for extracting information from these texts it is necessary that they be broken down into pieces so that the intention of the text is revealed completely without any ambiguity. In order to understand any Sanskrit text, this process of breaking down into individual words is necessary, and it is popularly known as \textit{sandhi-viccheda} (splitting of the joint text) in Sanskrit.\\

This process takes into account the morphological analyses of each of the split-parts obtained. As there is always a possibility for multiple morphological analyses even for individual words, considering only the morphological validation might result in enormous number of solutions for long sentences. So, syntactical accuracy is also measured to reduce the number of solutions. Even then, there is always a possibility for multiple solutions to remain, which cannot be resolved further without the semantic and contextual understanding of the sentence \cite{hellwig-tagger09:7}. Owing to this, we find that there is non-determinism right at the start of linguistic analysis \cite{huet-seg09}, since \textit{sandhi} splitting is the first step in the analysis of a Sanskrit sentence.\\

This kind of non-determinism is also found in languages like Chinese and Japanese, where the word boundaries are not indicated, and also in agglutinative languages like Turkish \cite{mittal-fst10:9}. In some of these languages like Thai \cite{harthai08}, most of the sentences have mere concatenation of words. Possible boundaries are predicted using the syllable information, and the process of segmentation starts with segmenting the syllable first, followed by the actual word segmentation. For Chinese though, their characters called \textit{hanzi} are easily identifiable, and the segmentation could be done by tagging \cite{xue-chin03}, or determining the word-internal positions using machine learning or deep learning algorithms (like what is done in \newcite{ji-emnlp18}). In the case of Vietnamese \cite{thang-vn08}, compound words are predominantly formed by semantic composition from 7000 syllables, which can also exist independently as separate words. This is similar to what can be observed in \textit{aluk sam\={a}sa} in Sanskrit, which are rare in occurrence. For languages like English, French and Spanish where the boundaries are specifically observed as delimiters like space, comma, semi-colon, full stop, etc., segmentation is done using these delimiters and is comparatively simple.

In all the above cases, we find that either there are delimiters to separate the words, or individual words are joined by concatenation which ultimately rests the segmentation process in the identification of boundaries. In the case of Sanskrit though, these kinds of words form a very small percentage. Rather, there is the euphony transformation that takes place at every word boundary. This transition can be generally stated as u|v $\rightarrow$ w, where u is the final part of the first word, v the first part of the next word, and w the resultant form after combining u and v. Here the parts may contain at the most two phonemes. The resultant w may contain additional phonemes or may have elisions, but never are more than two phonemes introduced. So this transition or \textit{sandhi} (external) occurs only at the phoneme level, and it does not require any other information regarding the individual words used.\footnote{In the case of internal \textit{sandhi} between preverbs and verbs, the lexical knowledge of the preverb is required. And in some compounds (like those denoting a \textit{sa\d{m}j\~{n}\={a}}), certain cases of retroflexion is permitted. But in this paper only the external \textit{sandhi} is considered.} But the reverse process of segmentation does require a morphological analyzer to validate the segments in a split.
And it is entirely up to the speaker or writer to perform these transitions or keep the words separated (called \textit{vivak\d{s}\={a}} - speaker's intention or desire). But in most of the texts and manuscripts, the \textit{sandhi} is done throughout the text. So, finding the split location alone will not be enough to segment the texts properly.\\

Having looked into some of the intricacies of \textit{sandhi} in Sanskrit, we can come up with a mechanical segmentation algorithm that splits a given text into all possible segments:
\begin{enumerate}
\item \label{item:first} Traverse through the input text and mark all possible split locations which could be found in the list of sandhied letters.\footnote{To get the list of sandhied letters, there is a list of \textit{s\={u}tras} or rules for the joining of letters, available in P\={a}\d{n}ini's \textit{A\d{s}\d{t}\={a}dhy\={a}y\={\i}} from which one can reverse analyze and obtain the list of sandhied letters.}
\item \label{item:second} When a sandhied letter is marked, then list all it's possible splits.
\item \label{item:third} Considering all the possible combinations of the words formed after each of these splits are allowed to join with the respective words (left word or right word), take each of the words, starting from the first word, to check for the morphological feasibility. Keep in mind that the words thus formed may also bypass the split locations, where they don't consider the split location present in between them.\footnote{For example - \textit{r\={a}m\={a}laya\d{h}} has split locations at 3 places - second, fourth (due to \textit{aka\d{h} savar\d{n}e d\={i}rgha\d{h} in \textit{A\d{s}\d{t}\={a}dhy\={a}y\={\i}} 6.1.101}) and sixth-seventh (due to \textit{eco'yav\={a}y\={a}va\d{h} in \textit{A\d{s}\d{t}\={a}dhy\={a}y\={\i}} 6.1.78}) letters. So, \textit{r\={a}} is one split word, as also \textit{r\={a}ma}, which bypasses the split location \textit{\={a}}. Similarly, we can find other split words also.}
\item \label{item:fourth} If the word is morphologically correct, then consider it as a valid split word and move on to the next split location, and do step \ref{item:third} until the last word of the sentence is reached. The sequence of words thus formed is the first solution.
If the word is not morphologically correct, move to step \ref{item:fifth}.
If all the words formed in a single split location, either on the left or on the right, or both, are not morphologically correct, then discard that split location and move to step \ref{item:third} for the next location.
\item \label{item:fifth} Check the words formed from the subsequent splits and continue with steps \ref{item:third} and \ref{item:fourth} to obtain other solutions.
\item \label{item:sixth} Trace back every split location, and perform step \ref{item:fifth}.
\item \label{item:seventh} In this way, get all the possible combinations of the split words.
\end{enumerate}

Although this mechanical process looks quite simple, the previously mentioned issues like non-determinism do prevail. And systems like the Sanskrit Reader in Sanskrit Heritage Engine come up with better ways to try to account these problems. The current paper tries to update these efforts. It is organized as follows: Section \ref{sec:current} gives the update on how the segmentation for Sanskrit has been dealt with in recent years. Section \ref{sec:segmenter} discusses the important features of The Sanskrit Heritage Engine's Reader. Section \ref{sec:issues} explains in detail the issues present in the Reader. The modifications needed to be done, and the implementation for this paper compose Section \ref{sec:modification}. It also quotes theoretically the reasons for these modifications and provides the probability function proposed in this paper. Section \ref{sec:methodology} describes the methodology of the implementation, and the results and observations are in Section \ref{sec:observations}.

\section{Current Methods} \label{sec:current}
Achieving the correct segmentation computationally is as much difficult as it is manually. A general approach would be the conversion of the mechanical \textit{sandhi} splitting process mentioned in Section \ref{sec:introduction} to a working algorithm, followed by checking the statistics available for the frequencies of the words and transitions. But there has been a lot of better research work, both rule-based and statistical, on computational \textit{sandhi} splitting in Sanskrit.

\newcite{huet-lex03}, as a part of the Sanskrit Heritage Engine, developed a Segmenter for Sanskrit texts using a Finite State Transducer. Two different segmenters were developed - one for internal \textit{sandhi}, which is deployed in the morphological analyser, and the other for external \textit{sandhi}. The current paper focuses on updating this external \textit{sandhi} segmenter.

\newcite{mittal-fst10:9} had used the Optimality Theory to derive a probabilistic method, and developed two methods to segment the input text \\(1) by augmenting the finite state transducer developed using OpenFst \cite{allau-fst07}, with \textit{sandhi} rules where the FST is used for the analysis of the morphology and is traversed for the segmentation, and \\(2) used optimality theory to validate all the possible segmentations.

\newcite{kumar-compound10:13} developed a compound processor where the segmentation for the compound words was done and used optimality theory with a different probabilistic method (discussed in section \ref{sec:modification}).

\newcite{charniak-s311:8} later modified the posterior probability function and also developed an algorithm based on Bayesian Word Segmentation methods with both unsupervised and supervised algorithms.

\newcite{pcrw16:10} proposed an approach combining the morphological features and word co-occurrence features from a manually tagged corpus from \newcite{hellwig-tagger09:7}, and took the segmentation problem as a query expansion problem and used Path Constrained Random Walk framework for selecting the nodes of the graph built with possible solutions from the input.

\newcite{vikas-segmenter18:16} built a word segmenter that uses a deep sequence to sequence model with attention to predict the correct solution. This is the state of art segmenter with precision and recall as 90.77 and 90.3, respectively.

IBM Research team \cite{rahul-seq210:15}, had built a Double Decoder RNN with attention as $seq2(seq)^2$, where they have emphasized finding the locations of the splits first, and then the finding of the split words. And they have the accuracy as 95\% and 79.5\% for finding the location of splits and the split sentence, respectively.

\newcite{hellwig-rcnn18:6} developed a segmenter using Character-level Recurrent and Convolutional Neural Networks, where they tokenize Sanskrit by jointly splitting compounds and resolving phonetic merges. The model does not require feature engineering or external linguistic resources. It works well with just the parallel versions of raw and segmented text.

\newcite{krishna-emnlp18} proposed a structured prediction framework that jointly solves the word segmentation and morphological tagging tasks in Sanskrit by using an energy based model which uses approaches generally employed in graph based parsing techniques.

\section{Heritage Segmenter} \label{sec:segmenter}
\begin{figure}
\centering
\includegraphics[scale=0.3]{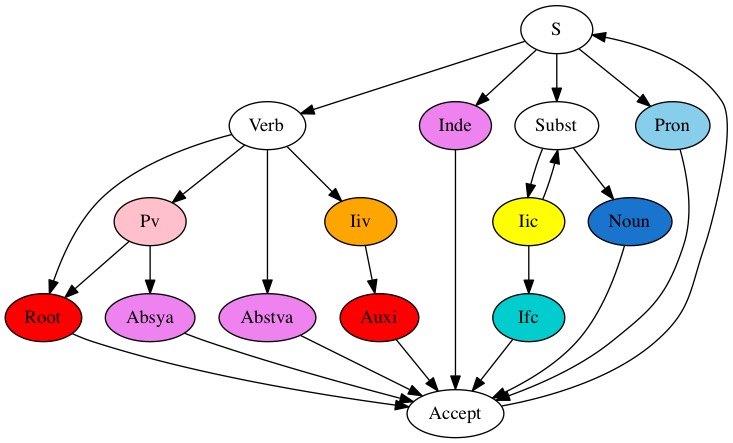}
\caption{A simplified lexical analyzer}
\label{fig:lex-analyzer}
\end{figure}

The Sanskrit Heritage Engine's Segmenter was chosen for further development, for three reasons - 
\begin{enumerate}
\item It is the best segmenter available online with source code available under GPL.
\item It uses a Finite State Transducer, and hence the segmentation is obtained in linear time.
\item It can produce all possible segmentations that one can arrive at, following P\={a}\d{n}ini's rules for \textit{sandhi}.
\end{enumerate}
It analyses the given input and produces the split based on three main factors:
\begin{enumerate}
\item Morphological feasibility: whether each of the words observed as a split is morphologically obtainable. 
\item Transition feasibility: whether every transition observed with each of the word is allowed.
\item Phase feasibility: whether the sequence of words have proper phase values. This is a constraint on the POS of a word. Although Sanskrit is a free word order language, there are certain syntactic constraints which govern the word formation, and the sequence of components within a word follows certain well defined syntax. The phase feasibility module takes care of this. Figure \ref{fig:lex-analyzer} shows a part of the lexical analyzer, developed by \newcite{goyal-reader13:12}, that portrays these phases like Iic, Inde, Noun, Root, etc.
\end{enumerate}

\begin{figure}
\centering
\includegraphics[width=0.5\linewidth]{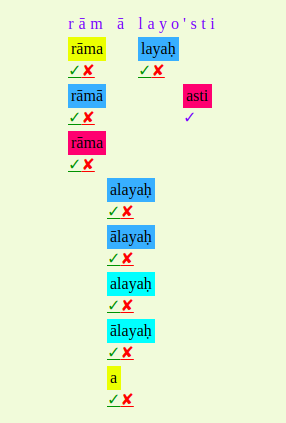}
\caption{The interface for choosing or rejecting the obtained split words for the example \textit{r\={a}m\={a}layo$'$sti}}
\label{fig:interface}
\end{figure}

Let us consider the sentence \textit{r\={a}m\={a}layo$'$sti}, as an example to understand these factors. It can be observed that there are twelve possible split solutions given in Table \ref{tab:solution}, from which, all the observed split words are shown in Figure \ref{fig:interface}.\\

\begin{table}[h]
\begin{center}
\begin{tabular}{|l|}\hline
Solutions\\\hline
r\={a}ma (iic) \={a}laya\d{h} (\={a}laya/\={a}li masc) asti\\\hline
r\={a}ma (iic) \={a}laya\d{h} (\={a}li fem) asti\\\hline
r\={a}ma (iic) alaya\d{h} (ali masc) asti\\\hline
r\={a}ma (iic) a (iic) laya\d{h} asti\\\hline
r\={a}ma (iic) alaya\d{h} (ali fem) asti\\\hline
r\={a}m\={a} (fem) laya\d{h} asti\\\hline
r\={a}m\={a} (fem) \={a}laya\d{h} (\={a}laya/\={a}li masc) asti\\\hline
r\={a}m\={a} (fem) alaya\d{h} (ali masc) asti\\\hline
r\={a}m\={a} (fem) a laya\d{h} asti\\\hline
r\={a}ma (r\={a}) \={a}laya\d{h} (\={a}laya/\={a}li masc) asti\\\hline
r\={a}ma (r\={a}) alaya\d{h} asti\\\hline
r\={a}ma (r\={a}) a (iic) laya\d{h} asti\\\hline
\end{tabular}
\caption{List of solutions for the sentence r\={a}m\={a}layo$'$sti}
\label{tab:solution}
\end{center}
\end{table}

Other possible words like \textit{r\={a}, m\={a}laya\d{h}}, etc. are not taken as proper splits because they do not form proper words according to the morphological analyzer present in the system. In this way, morphological feasibility is checked.
In the same example, we find that, at the last possible split location represented by \textit{o$'$}, we can split it as \textit{a\d{h}} and \textit{a} but not in any other way.\footnote{For the rules governing these transitions refer the \textit{A\d{s}\d{t}\={a}dhy\={a}y\={\i} s\={u}tra: atororaplut\={a}daplute (6.1.113) and ha\'{s}i ca (6.1.114)}} This is ensured by the transition feasibility module.

\begin{figure}
\centering
\includegraphics[width=0.5\linewidth]{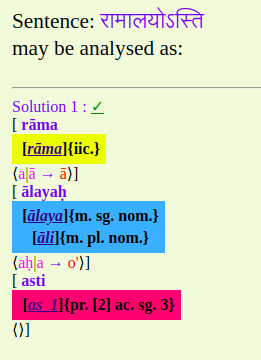}
\caption{The first solution for the sentence r\={a}m\={a}layo$'$sti}
\label{fig:sol1}
\end{figure}

The phase details like iic for \textit{r\={a}ma} or pr for \textit{asti}, etc. are displayed along with the words. These assignments of the phase information to the words and their analysis are the jobs of the phase feasibility module. To understand these phases, look at Figure \ref{fig:sol1} (the first solution for the sentence \textit{r\={a}m\={a}layo$'$sti}). \textit{r\={a}ma} is the first split and has the phase iic. \textit{\={a}laya\d{h}} is the second split with two morphological possibilities - \textit{\={a}laya} and \textit{\={a}li}. And the transition between the first two words is - a | \={a} $\rightarrow$ \={a}. The third split is \textit{asti} with root \textit{as} and phase pr. And the transition follows the equation - a\d{h} | a $\rightarrow$ o$'$. These transitions are taken care of by the transition feasibility module and the phases mentioned above are taken care of by the phase feasibility module.\\

According to \newcite{goyal-reader13:12}, sentences are formed by the image of the relation R (\textit{sandhi} rules) on Kleene closure of W*, of a regular set W of words (vocabulary of inflected words). The Sanskrit Heritage Reader accepts a candidate sentence w, and applies the inverted form of the relation R, thus producing a set of words - w1, w2, w3,.... And each of the individual words are valid according to the rules of morphology, and their combination makes some sense.\\

The methodology followed in the Segmenter proposed in \newcite{goyal-reader13:12} starts with using the finite state transducer for generating the chunks, instead of the traditional recursive method over the sentence employed in many \textit{sandhi} splitting tools. The FST considers the phases as important characteristics of the words. These phases correspond to a finite set of forms.

To understand how a word is obtained, let us first take a small example of how the substantival forms (\textit{subantas}) are obtained. A \textit{subanta} is analysed as a nominal stem followed by a suffix. The nominal stem can be either an underived stem or a derived stem. In case of a derived stem, the derivation of this stem is also provided by the segmenter.  A compound, for example, has a derived stem which contains a sequence of components followed by a nominal suffix. And three phases are present to represent the \textit{subantas}:
\begin{enumerate}
\item Noun, that contains declined forms of autonomous atomic substantive and adjective stems, from the lexicon
\item Ifc, non-autonomous and used as right-hand component of a compound
\item Iic contains bare stems of nouns to be used as left component
\end{enumerate}
This sequence of Subst $\rightarrow$ Noun $\rightarrow$ Accept, creates a noun word. And the sequence of Subst $\rightarrow$ Iic\textsuperscript{+} $\rightarrow$ Ifc $\rightarrow$ Accept, creates a compound word. These sequences can be observed in Figure \ref{fig:lex-analyzer}. In this way, forms from these phases are selected, and gluing them with \textit{sandhi} rules, a word is obtained. Considering all such possible phases in Sanskrit, an automation transition graph is formed and is used to traverse through to find the possible split locations and words together.\\

\section{Issues in Heritage Segmenter} \label{sec:issues}
The Segmenter is embedded in the Sanskrit Reader which displays all the outputs with the corresponding split word, it's phase and the transition involved with the subsequent word, except when the number of outputs is huge, in which case it shows only the summary. The Reader shows the distinction between words and phases based on verb, noun, iic, inde, etc, but not between some of the case-markers. So, there is inconsistency in disambiguation: sometimes the phase is used for pruning out certain solutions, but in some cases, it is not. For example, \textit{r\={a}movana\.{n}gacchati} produces the following 4 solutions:

{Solution 1 :} \\
{[ r\={a}ma\d{h} [r\={a}ma]{m. sg. nom.} ⟨a\d{h}|v → ov⟩]} \\
{[ vanam [vana]{n. sg. acc. | n. sg. nom.} ⟨m|g → \.{n}g⟩]} \\
{[ gacchati [gam]{pr. [1] ac. sg. 3} ⟨⟩]} \\

{Solution 2 :} \\
{[ r\={a}ma\d{h} [r\={a}ma]{m. sg. nom.} ⟨a\d{h}|v → ov⟩]} \\
{[ vanam [vana]{n. sg. acc. | n. sg. nom.} ⟨m|g → \.{n}g⟩]} \\
{[ gacchati [gacchat { ppr. [1] ac. }[gam]]{n. sg. loc. | m. sg. loc.} ⟨⟩]} \\

{Solution 3 :} \\
{[ r\={a}ma\d{h} [r\={a}\_1]{pr. [2] ac. pl. 1} ⟨a\d{h}|v → ov⟩]} \\present
{[ vanam [vana]{n. sg. acc. | n. sg. nom.} ⟨m|g → \.{n}g⟩]} \\
{[ gacchati [gam]{pr. [1] ac. sg. 3} ⟨⟩]} \\

{Solution 4 :} \\
{[ r\={a}ma\d{h} [r\={a}\_1]{pr. [2] ac. pl. 1} ⟨a\d{h}|v → ov⟩]} \\
{[ vanam [vana]{n. sg. acc. | n. sg. nom.} ⟨m|g → \.{n}g⟩]} \\
{[ gacchati [gacchat { ppr. [1] ac. }[gam]]{n. sg. loc. | m. sg. loc.} ⟨⟩]} \\

The segmenter provides segmentation and also does partial disambiguation. For example, \textit{r\={a}ma\d{h}} is ambiguous morphologically and the machine has correctly disambiguated the alternatives. We see the noun analysis of it in the first and second solutions, and the verbal analysis in third and fourth solutions. But we notice that the word \textit{vanam} which is ambiguous between two morphological analyses, one with nominative case marker and the other with accusative marker, is not disambiguated. \newcite{goyal-reader13:12} mention that the consideration of a word's similar declensions as different might result in more ambiguity, and the purpose of the segmentation is to find the morphologically apt words and hence they are taken as one. 

If we look at these four solutions, at the word level, all of them correspond to \textit{ r\={a}ma\d{h} vanam gacchati}. In order to decide the correct solution among the four, we need to do syntactico-semantic analyses that depend solely upon the linguistic or grammatical information in the sentence \cite{ambadepling13}.

\section{Proposed Modification} \label{sec:modification}

We notice that the use of phase information results in multiple solutions. In order to choose the correct solution among them, one needs to look beyond the word analysis and look at the possible relations between the words. This is the domain of the sentential parser. Only a sentential parser can decide which of the segmentations with phase information is the correct one. Thus we do not see any advantage of having the phase information.

And in the interface, the system is not uniform in resolving the ambiguities. It uses certain morphologically different phases under a single word, like \textit{vanam} in section \ref{sec:issues}. Additionally, in the options for selecting or rejecting the words, sometimes the depth of the graph goes so deep that, there is a chance to miss some solutions.

Here we would like to mention that some of the phase information is still relevant for segmentation. And this corresponds to the compounds. The phase information tells if something is a component of a compound or a standalone noun. There are a few phases such as iic, iif, etc. that we do not ignore. Barring these we ignore all other phases.\\

Therefore, we propose the following modifications in the segmenter:
\begin{enumerate}
\item Ignore the phase information that is irrelevant from segmentation point of view and merge the solutions that have the same word level segmentation.
\item Prioritize the solutions.
\end{enumerate}

This is similar to the intention in \newcite{vikas-segmenter18:16} where the morphological and other linguistic details are not obtained, but the segmentation problem is seen as an end in itself.

This is also similar to what \newcite{huet-seg09} did as an update for \newcite{gillon-compound10:14} to the compound analyzer where \newcite{gillon-compound10:14} uses the dependency structure to get the tree form consisting of all the parts of the compound word. And \newcite{huet-seg09} made the lexical analyzer to understand the compound as a right recursive linear structure of a sequence of components. This made sure that only the compound components are obtained, and not their relationship with each other. This helps in easier and faster segmentation, but the next level syntactic analysis cannot be done without the relationship information of the components.
Similarly, the same approach has been extended to all words, and not just compound words, and the phase details are not considered as valid parameters to distinguish solutions. Such solutions were termed duplicates and hence removed.

Once the duplicates are removed, prioritization needs to be done. Many probabilistic measures have been proposed in the past to prioritize the solutions. 

\newcite{mittal-fst10:9} calculated the weight for a specific split $s_j$ as
\begin{equation}
{W_s}_j = \frac{(\prod_{i=1}^{m-1} (\hat{P}(c_i) + \hat{P}(c_{i+1})) \times \hat{P}(r_i))}{m}
\end{equation}
where $\hat{P}(c_i)$ is the probability of the occurrence of the word $c_i$ in the corpus. $\hat{P}(r_i)$ is the probability of the occurrence of the rule $r_i$ in the corpus. And m is the number of individual components in the split $s_j$.

\newcite{kumar-compound10:13} uses the weight of the split $s_j$ as
\begin{equation}
{W_s}_j = \frac{(\prod_{i=1}^m \hat{P}(c_i)) \times (\prod_{i=1}^{m-1} \hat{P}(r_i))}{m}
\end{equation}

\newcite{charniak-s311:8} proposed a posterior probability function, $\hat{P}(s)$, the probability of generating the split $s = \langle c_1...c_m \rangle $, with $m$ splits, and rules $r = \langle r_1,...,r_{m-1} \rangle $ applied on the input, where
\begin{equation}
\hat{P}(s) = \hat{P}(c_1) \times \hat{P}(c_2 | c_1) \times \hat{P}(c_3 | c_2,c_1) \times ...\\
\end{equation}
\begin{equation}
\hat{P}(s) = \prod_{j=1}^m \hat{P}(c_j) \\
\end{equation}
$\hat{P}(c_1)$ is the probability of occurrence of the word $c_1$. $\hat{P}(c_2|c_1)$ is the probability of occurrence of the word $c_2$ given the occurrence of the word $c_1$, and so on.

\newcite{mittal-fst10:9} and \newcite{kumar-compound10:13} follow the GEN-CON-EVAL paradigm attributed to the Optimality Theory. This paper considers a similar approach but the probability function is taken as just the POP (product-of-products) of the word and transition probabilities of each of the solutions, discussed in section \ref{sec:methodology}.

And to prioritize the solutions, the following statistical data was added from the SHMT Corpus:\footnote{A corpus developed by the Sanskrit-Hindi Machine Translation (SHMT) Consortium under the funding from DeItY, Govt of India (2008-12). \url{http://sanskrit.uohyd.ac.in/scl/GOLD_DATA/tagged_data.html}}
\begin{itemize}[noitemsep,nolistsep]
\item \textit{sam\={a}sa} words with frequencies
\item \textit{sandhi} words with frequencies
\item \textit{sam\={a}sa} transition types with frequencies
\item \textit{sandhi} transition types with frequencies
\end{itemize}

\section{Methodology} \label{sec:methodology}
Every solution obtained after segmentation is checked for the two details viz. the word and the transition (that occurs at the end of the word due to the presence of the next word), along with the phase detail that is checked only for those which correspond to the components of a compound.
For every solution s, with output as 
\[s = \langle w_1.w_2....w_n \rangle\]
a confidence value, $C_i$, is obtained which is the product of the products of transition probablility (${P_t}_i$) and word probability (${P_w}_i$) for the word $w_i$,
\begin{equation}
C_i = \prod_{i=1}^n {P_w}_i \times {P_t}_i
\end{equation}
The confidence value is obtained as follows:
\begin{itemize}
\item For every split word $w_i$, it's phase is checked to know whether the obtained word forms a compound or not.
\item If it is a compound word, then it's corresponding frequency is obtained from compound words' statistical data, to calculate the word\_probability, $P(w_i)$
\item If it is not a compound word, then corresponding frequency is obtained from the \textit{sandhi} words' statistical data.
\item For every transition associated with the word, the transition's corresponding frequency is obtained from either the \textit{sam\={a}sa} transition data, or the \textit{sandhi} transition data, based on the phase of the word; to calculate the transition\_probability, $P(t_i)$.\footnote{If the frequency is not available for either the word or the transition, then it is assigned a default value of 1.}
\item The confidence value for the word, $w_i$ is thus obtained as the product of word\_probability and transition\_probability
$word\_probability \times transition\_probability$:
\begin{equation}
C_i = P_{w_i} \times P_{t_i}
\end{equation}
\item Finally the product of all such products was obtained for a single solution as the confidence value of the solution -
\begin{equation}
C_{total} = \prod_{i=1}^n P_{w_i} \times P_{t_i}
\end{equation}
\end{itemize}

The solutions are then sorted as decreasing order of confidence values and the duplicates are removed based on only the word splits. The remaining solutions are displayed along with their number and confidence values.

\section{Observations} \label{sec:observations}

The test data contained on the whole 21,127 short sandhied expressions, which were taken from various texts available at the SHMT corpus. This data was a parallel corpus of sandhied and unsandhied expressions. In case there are more than one segmentation possible, only one segmentation that was appropriate in the context where the sandhied expression was found is recorded.

\begin{table*}[t]
\begin{center}
\centering
\begin{tabular}{|l|c|c|c|c|}\hline
No. of & Old Segmenter & \% & Updated Segmenter & \% \\\hline
Input text & 21,127 & - & 21,127 & - \\\hline
Output text & 21,127 & - & 21,127 & - \\\hline
Correct sol & 19,494 & 92.27 & 19,494 & 92.27\\\hline
Correct sol in 1st & 10,432 & 53.51 & 17,403 & 89.27\\\hline
Correct sol in 2nd & 2,340 & 12.00 & 1,332 & 6.83\\\hline
Correct sol in 3rd & 1,874 & 9.61 & 429 & 2.20\\\hline
Correct sol in 4th & 937 & 4.8 & 164 & 0.84\\\hline
Correct sol in 5th & 703 & 3.6 & 73 & 0.37\\\hline
Correct sol in sol > 5th & 3,208 & 16.45 & 96 & 0.49\\\hline
Incorrect sol & 1,629 & 7.71 & 1,629 & 7.71\\\hline
Entries with 1 solution & 5,467 & 25.87 & 7,167 & 33.92\\\hline
Entries with 2 solutions & 3,320 & 15.71 & 4,002 & 18.94\\\hline
Entries with 3 solutions & 2,123 & 10.05 & 2,053 & 9.71\\\hline
\end{tabular}
\caption{A comparison of the performance of both the segmenters}
\label{tab:observe}
\end{center}
\end{table*}

The above data was fed to both the old and the modified segmenters. The results of the old segmenter were used as the baseline. A comparison was done on how the updated system performed with respect to the old system. The correct solution's position in the old segmenter was compared with the correct solution's position in the updated segmenter. Table \ref{tab:observe} summarizes the results.

The old segmenter was able to correctly produce the segmented form in 19,494 cases out of the 21,127 instances. Of these, 53.51\% of the solution was found to be in the first position, 12\% in second position, and 9.61\% in the third. All put together, 75.12\% of the correct solutions were found in the top three solutions.
Another important observation was that, the entire number of solutions taken all together was 2,40,942 for 21,127 test instances and the average number of solutions was 11.4 with the correct solution's position averaging at 4.71.

The modified segmenter was able to correctly produce the segmented form in 19,494 cases, same as the old segmenter. And 89.27\% of the solution was found to be in the first position, 6.83\% in the second position, and 2.2\% in the third. All put together, 98.3\% of the correct solutions were found in the top three solutions. This has an increase of 23.18\% from the existing system.
Also, the entire number of solutions taken all together was 1,46,610 for 21,127 test instances, having a drastic reduction of 94,332 solutions. The average number of solutions was 6.94, with the correct solution's position averaging at 1.18.

It can be noted that the overall Recall was 0.92270554267 for both the machines. Since only the statistics have been altered, the new system doesn't provide new solutions. Rather, it has increased the chances of getting the solution at the top three by 23.18\%.

As we observe, the updated system reduces the total amount of solutions and brings up the most likely solutions. Also, we have more than 90\% recall in both the cases. The missed out instances were either due to morphological unavailability or owing to the failure of the engine. Once the morphological analyzer is updated, there will definitely be a boost in the efficiency.

\section{Conclusion} \label{sec:conclusion}

There are a few observations to be noted. First, by just using the POP (product of products) of the word and transition probabilities, we are able to obtain 98\% precision. With better probabilities, we will definitely have better results. Second, this system can now be used to mechanically split the continuous texts like \textit{Sa\d{m}hita-P\={a}\d{t}ha} of the Vedas or any other classical text to obtain the corresponding \textit{Pada-P\={a}\d{t}ha}, which may be manually checked for correctness. Third, for mere segmentation, the phase distinctions were ignored, and the obtained solutions were prioritized. As stated earlier in the previous sections, to proceed to the next stage of parsing or disambiguation, we need more than just the split words. Thus this could be a proper base for working on how the available segmented words, along with the phase details, may be used for further stages of analysis.

\nocite{*}

\bibliography{acl2019}
\bibliographystyle{acl_natbib}

\end{document}